\documentclass[10pt,twocolumn,letterpaper]{article}

\usepackage{cvpr}              

\usepackage{xcolor}

\usepackage{booktabs}
\usepackage{multirow}
\usepackage{multirow}
\usepackage{float}

\usepackage{dblfloatfix}
\usepackage{cuted}

\definecolor{cvprblue}{rgb}{0.21,0.49,0.74}
\usepackage[pagebackref,breaklinks,colorlinks,allcolors=cvprblue]{hyperref}

\def\paperID{23} 
\def\confName{CVPR}
\def\confYear{2026}

\title{Pretraining Objective Matters in Extreme Low-Data FGVC: \\A Backbone-Controlled Study}

\author{
Alexander Hackett\\
Santa Clara University\\
{\tt\small ahackett@scu.edu}
\and
Srikanth Thudumu\\
IAAIR\\
{\tt\small srikanth@iaair.ai}
\and
Ginny Fisher\\
IAAIR\\
{\tt\small ginny.fisher@iaair.ai}
\and
Jason Fisher\\
IAAIR\\
{\tt\small jason@iaair.ai}
}

\begin{document}
\maketitle
\begin{abstract}
Extreme low-data fine-grained classification is common in expert domains where labeling is expensive, yet practitioners still need principled guidance for selecting pretrained encoders. We study emerald inclusion grading with a custom dataset of labeled images across three classes and ask: under matched backbone capacity, how does pretraining objective affect downstream representation quality? We compare four frozen ViT-B/16 encoders trained with supervised classification, contrastive learning (SigLIP2), masked reconstruction (MAE), and self-distillation (DINOv3), and evaluate them with leave-one-out cross-validation using linear and nonlinear probes. To control statistical noise in the low-$N$ regime, we use permutation testing ($N_{perm}=1000$) on macro one-vs-rest AUC. Supervised and contrastive encoders provide the strongest linear separability (logistic AUC: 0.768 and 0.735; SVM AUC: 0.739 and 0.697), while MAE improves under nonlinear probes (XGBoost AUC: 0.713). We find that DINOv3 underperforms across probe families in this domain. These results support a practical recommendation for extreme low-data FGVC: prioritize margin-enforcing pretraining objectives when data scarcity restricts probing to linear decision rules, and consider reconstruction-style encoders when nonlinear classifiers are feasible given dataset constraints.
\end{abstract}    
\section{Introduction}
\label{sec:intro}

Fine-grained visual categorization (FGVC) often operates in data regimes where labels are scarce, costly, and tied to domain expertise. Emerald inclusion grading is one such case: certified gemologists assign categories (eye-clean, moderately included, heavily included) from subtle internal visual cues, but assembling large annotated datasets is impractical. In practice, each label requires inclusion assessment under magnification by a trained gemologist (typically GIA-certified), making annotation slow and expensive and limiting feasible dataset scale. Our benchmark contains 37 images (class counts 21/9/7), making end-to-end fine-tuning unstable and placing emphasis on frozen-representation quality.

This paper asks a targeted question for this setting: \emph{when backbone architecture is controlled, how much does the pretraining objective determine downstream separability?} We standardize encoder capacity by comparing ViT-B/16 models only (\textasciitilde86M parameters) across four objective families: supervised classification on ImageNet \cite{russakovsky2015imagenet}, contrastive learning (SigLIP2) \cite{tschannen2025siglip2}, masked reconstruction (MAE) \cite{he2022mae}, and self-distillation (DINO family) \cite{caron2021dino,simeoni2025dinov3}. Recent empirical comparisons of supervised and contrastive pretraining under robustness-oriented evaluation further motivate this question \cite{vishniakov2024convnet}.

Our evaluation protocol follows linear probing as a representation diagnostic \cite{alain2016probes}: we keep encoders frozen, train lightweight downstream probes, and evaluate via leave-one-out cross-validation (LOOCV). Because $N=37$ makes split variance large, we complement point estimates with permutation testing ($N_{perm}=1000$) on macro one-vs-rest AUC \cite{ojala2010permutation}. We report linear probes (logistic regression, linear SVM) as primary diagnostics and nonlinear probes (XGBoost, Random Forest) as secondary stress tests.

We use this protocol to report application-grounded findings for emerald grading under real annotation constraints, focusing on which pretrained representations are most usable under linear probing \cite{wang2020alignmentuniformity,zhang2023linearseparability}.

\begin{figure*}[t]
    \centering
    \begin{subfigure}[b]{0.32\linewidth}
        \centering
        \includegraphics[width=\linewidth]{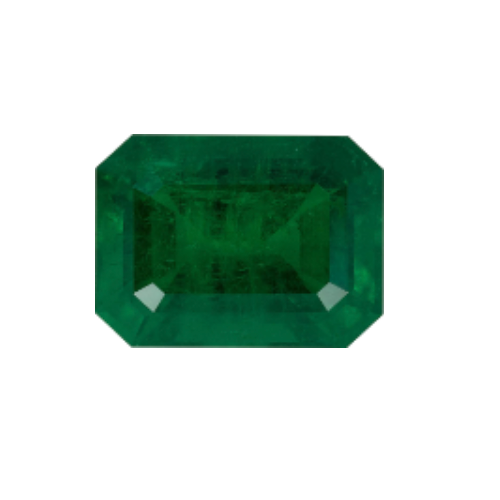}
        \caption{Eye-clean}
    \end{subfigure}
    \hfill
    \begin{subfigure}[b]{0.32\linewidth}
        \centering
        \includegraphics[width=\linewidth]{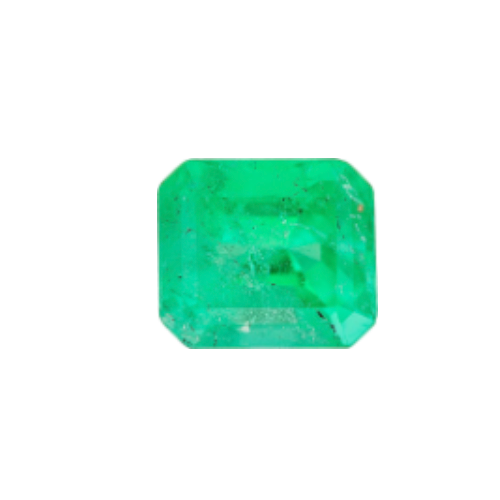}
        \caption{Moderate}
    \end{subfigure}
    \hfill
    \begin{subfigure}[b]{0.32\linewidth}
        \centering
        \includegraphics[width=\linewidth]{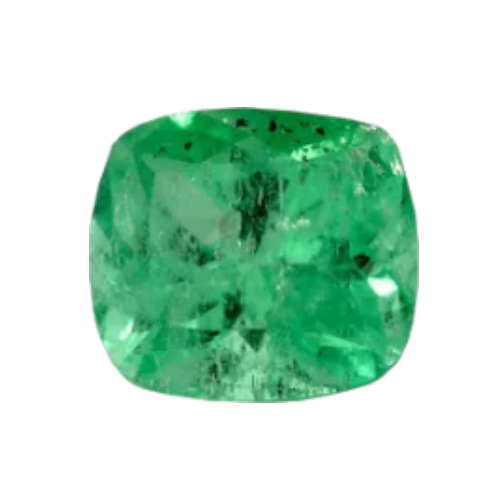}
        \caption{Heavy}
    \end{subfigure}
    \caption{Representative specimens from each clarity grade. (a) Eye-clean: saturated green, continuous color, no eye-visible inclusions. (b) Moderate: dispersed dark jardin that does not substantially reduce transparency. (c) Heavy: concentrated jardin deposits with turbid zones and loss of color continuity.}
    \label{fig:clarity_examples}
\end{figure*}

\paragraph{Contributions.}
\begin{itemize}
    \item We present a backbone-controlled comparison of four pretraining objective families for extreme low-data FGVC using matched ViT-B/16 encoders.
    \item We provide a statistically grounded ranking protocol using LOOCV with permutation testing ($N_{perm}=1000$) over macro AUC.
    \item We show that margin-enforcing objectives (supervised, contrastive) yield stronger linear separability, while reconstruction signal (MAE) is more recoverable with nonlinear probes.
    \item We introduce a controlled perturbation diagnostic in which synthetic inclusions are manually added to eye-clean specimens, testing whether encoder representations respond to inclusion-relevant cues rather than global image properties. We find that encoders trained on contrastive objectives show the strongest perturbation response, while encoders pretrained on reconstruction-style objectives show near-zero linear sensitivity.
\end{itemize}

\section{Method}
\label{sec:method}

\subsection{Task and data}
We study three-class emerald inclusion grading on a labeled set of RGB emerald images on a uniformly white backdrop. The class distribution is imbalanced (21/9/7), so we report macro metrics throughout.

Emerald clarity assessment is inclusion-severity grading, not inclusion detection. Following standard gemological trade terminology documented by GIA \cite{gia2026emeraldquality}, we use three practical classes: \emph{eye-clean} (no inclusions visible to the unaided eye), \emph{moderately included} (visible jardin/inclusions that do not severely reduce transparency), and \emph{heavily included} (inclusions that materially degrade transparency and clarity). Labels were assigned by a GIA-certified gemologist under magnified inspection. Representative specimens for each class are shown in Fig.~\ref{fig:clarity_examples}.

\subsection{Backbone-controlled encoder set}
To ensure there are no confounds across architecture and scale, all learned encoders use a ViT-B/16 backbone (\textasciitilde86M parameters, 768-dim embeddings). \autoref{tab:encoder_set} summarizes the compared objective families and pretraining sources.

\begin{table}[t]
    \centering
    \scriptsize
    \caption{Backbone-controlled encoder set used for probing.}
    \label{tab:encoder_set}
    \resizebox{\linewidth}{!}{%
    \begin{tabular}{llll}
        \toprule
        Encoder & Objective & Params & Pretrain data \\
        \midrule
        Supervised ViT-B/16 & Supervised Classification & \textasciitilde86M & ImageNet \\
        SigLIP2-B/16 & Contrastive Language-Image & \textasciitilde86M & WebLI \\
        MAE-B/16 & Masked Reconstruction & \textasciitilde86M & ImageNet \\
        DINOv3-B/16 & Self-Distillation & \textasciitilde86M & LVD-style corpus \\
        \bottomrule
    \end{tabular}
    }
\end{table}

As spurious-separability controls, we include (i) random Gaussian embeddings, denoted $\mathcal{N}(0,I)$ with $d=768$, and (ii) a classical handcrafted-feature baseline adapted from Pe\~{n}a et~al.~\cite{pena2022emerald}: HSV color statistics (mean and standard deviation of saturation and value channels), GLCM texture descriptors (homogeneity and entropy at two orientations), and Bhattacharyya histogram distances to per-class reference specimens (14-dimensional total).

\subsection{Frozen probing protocol}
For each encoder, we extract one embedding per image and train downstream probes with encoder weights frozen. Evaluation uses LOOCV so each fold trains on 36 samples and tests on one held-out sample.

\textbf{Primary probes (linear):} multinomial logistic regression and linear SVM. \\
\textbf{Secondary probes (nonlinear):} XGBoost and Random Forest.

We report macro Accuracy, macro F1, and macro one-vs-rest AUC. Our primary ranking statistic is macro AUC.

\subsection{Permutation testing}
For each encoder--probe pair, we estimate significance with a label-permutation test \cite{ojala2010permutation}: labels are shuffled 1000 times, the full LOOCV AUC is recomputed, and the $p$-value is the fraction of null AUCs greater than or equal to the observed AUC. This reduces over-interpretation risk in high-dimensional, low-sample settings.

\begin{figure*}[!t]
    \centering
    \scriptsize
    \captionof{table}{LOOCV performance with permutation testing ($N_{perm}=1000$). Primary comparisons are among matched ViT-B/16 encoders: Supervised ViT, SigLIP2, MAE, and DINOv3. $\mathcal{N}(0,I)$ denotes random Gaussian embeddings ($d=768$). Permutation $p$-values are not reported for the $\mathcal{N}(0,I)$ control, whose embeddings are label-independent by construction. Lower permutation $p$ indicates stronger evidence beyond chance.}
    \label{tab:main_results}
    \resizebox{\textwidth}{!}{%
    \begin{tabular}{lcccccccccccccccc}
        \toprule
        & \multicolumn{4}{c}{Logistic} & \multicolumn{4}{c}{Linear SVM} & \multicolumn{4}{c}{XGBoost} & \multicolumn{4}{c}{Random Forest} \\
        \cmidrule(lr){2-5} \cmidrule(lr){6-9} \cmidrule(lr){10-13} \cmidrule(lr){14-17}
        Encoder & Acc & AUC & F1 & $p$ & Acc & AUC & F1 & $p$ & Acc & AUC & F1 & $p$ & Acc & AUC & F1 & $p$ \\
        \midrule
        $\mathcal{N}(0,I)$ ($d=768$) & 0.568 & 0.474 & 0.299 & -- & 0.541 & 0.434 & 0.234 & -- & 0.676 & 0.689 & 0.565 & -- & 0.514 & 0.403 & 0.283 & -- \\
        Supervised ViT & \textbf{0.595} & \textbf{0.768} & \textbf{0.512} & \textbf{0.003} & \textbf{0.568} & \textbf{0.739} & \textbf{0.511} & \textbf{0.005} & 0.541 & 0.604 & 0.455 & 0.121 & \textbf{0.676} & \textbf{0.713} & \textbf{0.575} & \textbf{0.004} \\
        DINOv3 & 0.432 & 0.625 & 0.307 & 0.079 & 0.432 & 0.530 & 0.307 & 0.277 & 0.486 & 0.470 & 0.362 & 0.439 & 0.486 & 0.497 & 0.360 & 0.273 \\
        SigLIP2 & 0.486 & 0.735 & 0.411 & 0.011 & \textbf{0.568} & 0.697 & 0.439 & 0.018 & \textbf{0.622} & 0.706 & \textbf{0.524} & \textbf{0.021} & 0.486 & 0.596 & 0.413 & 0.065 \\
        MAE & 0.486 & 0.636 & 0.361 & 0.053 & 0.486 & 0.551 & 0.359 & 0.209 & 0.568 & \textbf{0.713} & 0.465 & 0.023 & 0.568 & 0.603 & 0.441 & 0.018 \\
        Classical features & 0.351 & 0.560 & 0.323 & 0.193 & 0.324 & 0.548 & 0.308 & 0.250 & 0.405 & 0.494 & 0.361 & 0.365 & 0.297 & 0.447 & 0.270 & 0.535 \\
        \bottomrule
    \end{tabular}%
    }
\end{figure*}

\section{Results}
\label{sec:results}

\subsection{Primary diagnostic: linear separability}
Table~\ref{tab:main_results} shows that linear probes strongly favor supervised and contrastive pretraining. For logistic regression, Supervised ViT achieves the highest AUC (0.768, $p=0.003$), followed by SigLIP2 (0.735, $p=0.011$). The same ordering appears under linear SVM (0.739, $p=0.005$ for Supervised ViT; 0.697, $p=0.018$ for SigLIP2). MAE and DINOv3 do not reach statistical significance under linear probes.
The $\mathcal{N}(0,I)$ control ($d=768$) remains weak under linear probes (AUC 0.474 logistic, 0.434 SVM), while its stronger XGBoost score (AUC 0.689) illustrates why nonlinear probe outputs must be interpreted cautiously in this low-$N$ regime. Classical features remain consistently weak across probe families.

\subsection{Nonlinear probes and HDLSS caution}
Nonlinear probes reveal a different pattern. Under XGBoost, MAE reaches AUC 0.713 ($p=0.023$), outperforming Supervised ViT and slightly exceeding SigLIP2 in AUC. This suggests MAE embeddings contain usable class information that is less linearly accessible. Random Forest again favors Supervised ViT (AUC 0.713, $p=0.004$), while MAE remains competitive (AUC 0.603, $p=0.018$).

Together, linear and nonlinear results imply a practical distinction: some objectives expose information directly to linear decision boundaries, whereas others preserve information that requires nonlinear extraction.

\subsection{Perturbation sensitivity}
To test whether encoder representations respond to
inclusion-relevant visual cues rather than global image
statistics (e.g.\ stone shape, background, overall color),
we introduce a controlled perturbation diagnostic. For each
eye-clean specimen ($N{=}21$), we manually edited the image
using Adobe Photoshop to insert dark deposits within the
 stone interior, producing a matched pair in which only 
 the inclusion-relevant region has changed.
Edits preserved each stone's outline, lighting, and
background so that encoder response is attributable to
interior cues alone; Fig.~2 shows a representative pair.

\begin{figure}[H]
    \centering
    \includegraphics[width=0.48\linewidth]{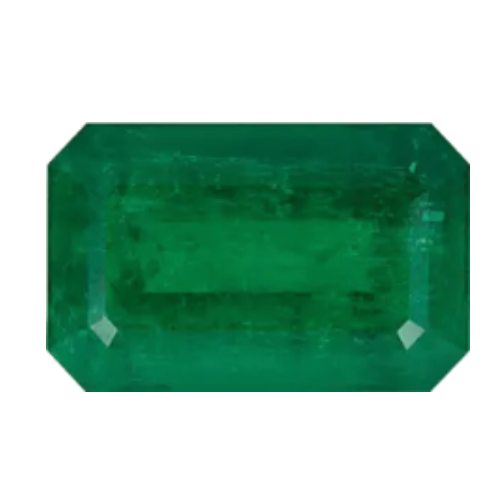}\hfill
    \includegraphics[width=0.48\linewidth]{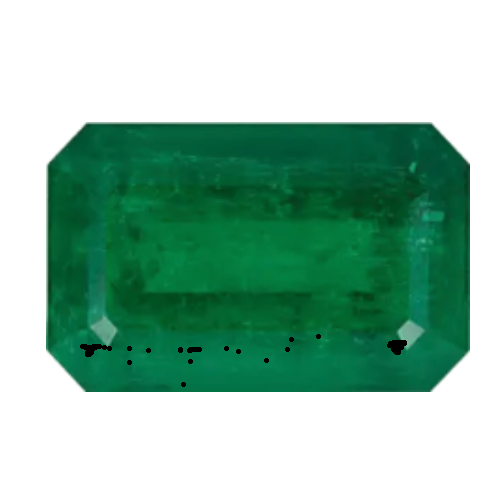}
    \caption{Representative clean/perturbed pair. Left: original eye-clean specimen. Right: manual edit with dark deposits and diffuse grey streaks added to the stone interior. All other image properties are preserved.}
    \label{fig:adversarial_pair}
\end{figure}

For each encoder we define the \emph{eye-clean margin}
under the frozen logistic probe as
\begin{equation}
  m(x) \;=\; z_0(x) \;-\; \max_{k \neq 0}\, z_k(x),
\end{equation}
where $\mathbf{z}(x) \in \mathbb{R}^3$ is the logit vector
and class~0 is eye-clean. The \emph{perturbation margin
drop} for specimen $i$ is
$\Delta m_i = m(x_i) - m(\tilde{x}_i)$,
where $\tilde{x}_i$ is the perturbed version. A positive
$\Delta m_i$ indicates that the perturbation narrowed the
logit gap between the eye-clean class and its strongest
competitor, i.e.\ the perturbed embedding lies closer to
the decision boundary. We report $\overline{\Delta m}$, its
standard deviation, and the \emph{reclassification rate}:
the fraction of specimens whose predicted class switches
away from eye-clean after perturbation (Table~3).

\begin{table}[H]
    \centering
    \scriptsize
    \caption{Perturbation sensitivity on edited eye-clean specimens ($N{=}21$). Larger margin drop and higher reclassification rate indicate stronger response to synthetic inclusion cues.}
    \label{tab:perturbation}
    \begin{tabular}{lccc}
        \toprule
        Encoder & Avg. margin drop & Std. margin drop & Reclass rate \\
        \midrule
        Supervised ViT & 3.13 & 2.80 & 19.0\% \\
        SigLIP2 & \textbf{4.61} & 2.46 & \textbf{38.1\%} \\
        MAE & 0.02 & 1.16 & 0.0\% \\
        DINOv3 & 1.91 & 1.29 & 4.8\% \\
        \bottomrule
    \end{tabular}
\end{table}

SigLIP2 exhibits the strongest perturbation response
($\overline{\Delta m} = 4.61$, reclass.\ 38.1\%), despite
ranking second to Supervised ViT under linear probes. This
suggests contrastive pretraining produces representations
that are sensitive to local interior disruptions. Supervised
ViT responds moderately ($\overline{\Delta m} = 3.13$,
reclass.\ 19.0\%). MAE shows near-zero mean response
($\overline{\Delta m} = 0.02$) but high specimen-level
variance ($\sigma = 1.16$), suggesting inconsistent rather
than uniformly absent sensitivity to these perturbations.
DINOv3 responds
weakly (reclass.\ 4.8\%), reinforcing that objective family
materially affects perturbation sensitivity even under
matched backbones.

\section{Discussion}
\label{sec:discussion}

The results support three takeaways for extreme low-data FGVC.

\textbf{(1) Linear accessibility tracks objective geometry, not just information content.} Under fixed ViT-B/16 capacity, margin-enforcing objectives are more directly readable by linear probes, while other objectives can retain signal that is less linearly exposed.

\textbf{(2) MAE is not uniformly weak; it is linearly underexposed.} MAE underperforms under linear probes but recovers under nonlinear ones: its Random Forest AUC rises to 0.603 ($p = 0.018$), marginally outperforming SigLIP2 (AUC 0.596, $p = 0.065$) and significantly outperforming DINOv3 (AUC 0.497, $p = 0.273$). Its XGBoost AUC reaches 0.713 ($p = 0.023$), though the latter should be interpreted cautiously given the $\mathcal{N}(0,I)$ control ($d=768$) XGBoost AUC of 0.689. This gap between linear and nonlinear probe performance suggests that masked reconstruction preserves class-relevant structure in a geometry that linear decision boundaries cannot access. The perturbation analysis reinforces this interpretation: MAE's near-zero margin drop ($\overline{\Delta m} = 0.02$) indicates that its representations do not linearly encode inclusion severity, even though the information may reside in higher-order feature interactions that nonlinear tree-based probes exploit. For practitioners, this implies MAE remains a viable encoder when nonlinear downstream classifiers are accessible, but should be avoided when linear-probe reliability is the deployment constraint.

\textbf{XGBoost caution with random high-dimensional controls.} The $\mathcal{N}(0,I)$ control ($d=768$) reaches a relatively high XGBoost AUC despite lacking semantic structure, while remaining weak under linear probes. This reinforces our use of linear probe significance as the primary diagnostic and motivates treating nonlinear gains as secondary, hypothesis-generating evidence in extreme low-data settings.

\textbf{(3) DINOv3 features are weakest in this domain setup.} DINOv3 underperforms across probe families, suggesting that self-distillation invariances learned at pretraining scale do not transfer effectively to this fine-grained, low-$N$ benchmark without further adaptation. One structural difference that may explain DINOv3's underperformance relative to SigLIP2 is that the SigLIP2 objective operates directly on dot products between L2-normalized embeddings, so pairwise repulsion is enforced in the same geometry that linear probes read. DINOv3's self-distillation minimizes cross-entropy over softmax-normalized distributions, where the mapping from embedding space to probability simplex is many-to-one. Large distributional divergence does not guarantee geometric separation in the underlying representation. This may explain why DINOv3's features are less linearly accessible despite strong performance on in-distribution benchmarks.

These observations are consistent with the intuition that margin-enforcing objectives can improve linear accessibility \cite{wang2020alignmentuniformity}, while broader architecture studies indicate separability potential is not solely a backbone-family effect \cite{zhang2023linearseparability}.

\paragraph{Limitations.}
This is a single-domain study with only 37 labeled images. Although backbones are controlled, pretraining data sources are not matched, so objective and corpus effects remain partially entangled. We therefore frame conclusions as application-scoped evidence and as guidance for low-data model selection, not as a universal ranking of representation learning paradigms.
\section{Conclusion}
\label{sec:conclusion}

We presented a backbone-controlled low-data FGVC study for emerald inclusion grading. Using matched ViT-B/16 encoders, LOOCV, and permutation-tested macro AUC ($N_{perm}=1000$), we find that supervised and contrastive objectives yield the most linearly separable embeddings, while MAE becomes competitive under nonlinear probes. A controlled perturbation diagnostic reinforces this: when synthetic inclusions are introduced to eye-clean specimens, contrastive encoders produce the largest logit margin drop and change in decision under the frozen linear probe, suggesting their representations most reliably encode inclusion-relevant signal in linearly accessible geometry. DINOv3 representations remain comparatively weak across probe types and perturbation sensitivity.

When data scarcity restricts practitioners to linear probes, margin-enforcing encoders (supervised, contrastive) should be preferred. When dataset size can support nonlinear classifiers, reconstruction-pretrained encoders become competitive.
\section*{Acknowledgements}
We thank Aisha Sartaj and Mahule Roy for their early feedback and general discussions during the initial planning phase of this work.
{
    \small
    \bibliographystyle{ieeenat_fullname}
    \bibliography{main}
}

\end{document}